\documentclass[letterpaper, 10 pt, conference]{ieeeconf}  

\IEEEoverridecommandlockouts                              

\usepackage{graphicx}
\usepackage{amsmath} 
\usepackage{amssymb}  
\usepackage{lipsum}
\usepackage{mathtools}
\usepackage{siunitx}
\usepackage{wasysym}
\usepackage{array}
\newcolumntype{C}[1]{>{\centering\let\newline\\\arraybackslash\hspace{0pt}}m{#1}}

\newcommand\myeq{\stackrel{\mathclap{\scriptsize\mbox{def}}}{=}}

\title{\LARGE \bf
Design and Workspace Characterisation of Malleable Robots
}

\author{Angus B. Clark,~\IEEEmembership{Student Member, IEEE} and Nicolas Rojas,~\IEEEmembership{Member, IEEE}%
\thanks{Angus B. Clark and Nicolas Rojas are with the REDS Lab, Dyson School of Design Engineering, Imperial College London, 25 Exhibition Road, London, SW7 2DB, UK
{\tt\small (a.clark17, n.rojas)@imperial.ac.uk}}%
}

\begin{document}

\maketitle
\thispagestyle{empty}
\pagestyle{empty}

\begin{abstract}
For the majority of tasks performed by traditional serial robot arms, such as bin picking or pick and place, only two or three degrees of freedom (DOF) are required for motion; however, by augmenting the number of degrees of freedom, further dexterity of robot arms for multiple tasks can be achieved. Instead of increasing the number of joints of a robot to improve flexibility and adaptation, which increases control complexity, weight, and cost of the overall system, malleable robots utilise a variable stiffness link between joints allowing the relative positioning of the revolute pairs at each end of the link to vary, thus enabling a low DOF serial robot to adapt across tasks by varying its workspace. In this paper, we present the design and prototyping of a 2-DOF malleable robot, calculate the general equation of its workspace using a parameterisation based on distance geometry---suitable for robot arms of variable topology, and characterise the workspace categories that the end effector of the robot can trace via reconfiguration. Through the design and construction of the malleable robot we explore design considerations, and demonstrate the viability of the overall concept. By using motion tracking on the physical robot, we show examples of the infinite number of workspaces that the introduced 2-DOF malleable robot can achieve.
\end{abstract}

\section{Introduction}

When considering the design of any robot manipulator, the first and foremost design consideration is the number of degrees of freedom (DOF) the robot will have. To determine the DOF, typically the tasks for the robot are considered, determining the requirements of the design \cite{al-dois-task}. For the majority of tasks performed by serial robots, only a low DOF is required \cite{yang-task}. However, determining this, and thus the required parameters of the robot, can be difficult to predict \cite{chocron-algorithms}. Serial robot arms are therefore either designed to be hyper-redundant with more DOF than required, and therefore expensive, or are designed with the exact DOF required, which can limit their scope of application \cite{cohen-conceptual}.

Soft robotic manipulators have provided one solution to the issue of requiring high DOF for achieving high dexterity, with the development of continuously bending manipulators (continuum robots) \cite{walker-continuous, marchese-design}. The natural conformity and bending of soft manipulators allows for a high adaptability across tasks \cite{rus-design}. However, accurate motion planning and control, as well as high structural strength, have proved difficult to achieve \cite{renda-3d}. Variable stiffness technologies have recently been implemented in these robots to improve their strength while maintaining their flexibility \cite{ranzani-modular, kim-soft, cianchetti-soft}.

The design of robot mechanisms that are capable of changing the mode of motion or the number of DOF of the robot have been called reconfigurable \cite{lopez-reconfigurable}. The significant majority of proposed research regarding reconfigurable mechanisms has focused on the design of modular robot arms, that are capable of providing system flexibility \cite{chocron-evolutionary, baca-heterogeneous, bi-concurrent}. However, these designs are still limited by the complexity of their control and motion planning. Development of serial robot arms that are capable of achieving this flexibility while maintaining low complexity has been limited. We propose to solve this problem using malleable robots, which can be also catalogued as reconfigurable robot arms. Malleable Robots are defined as reduced-DOF serial arms of changeable geometry \cite{clark-stiffness}, whereby the integration of a variable stiffness continuously bending link between joints allows the relative positioning of revolute pairs to vary, producing a variable robot topology, while the reduced DOF maintains the simplicity of the system.

\begin{figure}[t!]
    \centering
    \includegraphics[width=\columnwidth]{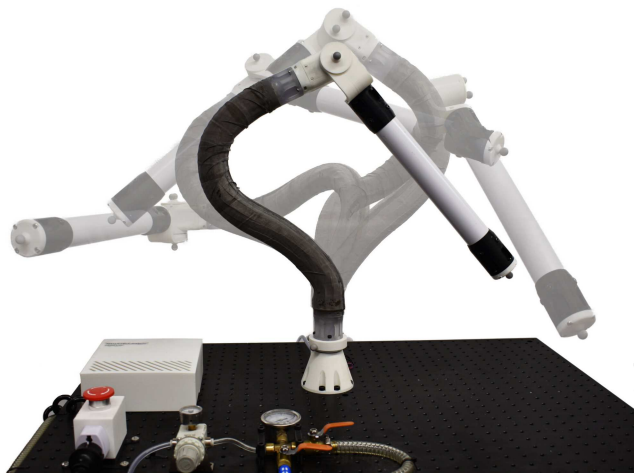}
    \caption{The developed two-degree-of-freedom (DOF) malleable robot arm, showing various topology configurations it can achieve. A PUMA-like configuration is shown in foreground.}
    \label{malleablerobot}
\end{figure}

The inherent non-fixed robot geometry of malleable robots allows the generation of an infinite number of workspaces for adapting the system to different tasks. The workspace of a robot arm is defined as the region (or surface/volume) within which every point can be reached by the end effector, and is one of the most important specifications for both robot designers and users \cite{tsai-algorithm}. We can either compute the workspace given the structure (analysis), or alternatively determine the robot structure from a desired workspace (synthesis) \cite{gupta-design}. For the case of malleable robots, traditional strategies for the computation of workspaces based on the attachment of reference frames to the robot joints, such as those that make use of the Denavit-Hartenberg convention \cite{kung2005development, li2011design}, cannot be employed since both link dimensions and the relative orientation of the joints can change. An alternative is to perform the workspace analysis using screw theory \cite{coppola20146, xie2015design} or distance geometry \cite{rojas-thesis} as in these approaches the parameterisation does not depend on relative angles and distances between joint reference frames. We make use of a distance-geometry-based method herein as the technique has been shown to simplify the computation of the workspace equation of complex mechanisms \cite{rojas-coupler, rojas-peaucellier}.

We introduce a first-of-its-kind malleable robot, as shown in Fig.~\ref{malleablerobot}, consisting of two rotary joints, a malleable link of variable stiffness with structural spine designed for collaborative extrinsic reconfiguration \cite{clark-stiffness, clark-assessing}, and a rigid link connecting the second joint to the end effector. By lowering the stiffness of the malleable link via vacuum elimination, this can be reshaped by hand as desired. Then, by stiffening the link again using negative pressure, the joints are fixed in place, forming a new robot topology. To the best of the authors' knowledge, no workspace calculation or analysis has yet been performed for malleable robots, and this work is the first demonstration of the viability and capabilities of this technology, presenting physical examples of the infinite number of workspaces that a 2-DOF malleable robot can achieve. 

The rest of this paper is organised as follows. Firstly, in section II, the distance-based parameters and main set of formulae defining the malleable robot workspace are identified, and then the symbolic equation of the workspace surface traced by the end effector of a 2-DOF malleable robot is obtained, along with a presentation of its workspace categories. In section III, we discuss the design of the malleable robot, with considerations of specific aspects of this, such as the malleable link and joints. In section IV and V, we present and comment on the workspaces generated by the malleable robot prototype. Finally, we conclude in section VI.

\section{Workspace Definition}

\begin{figure}[t!]
    \centering
    \includegraphics[width=0.9\columnwidth]{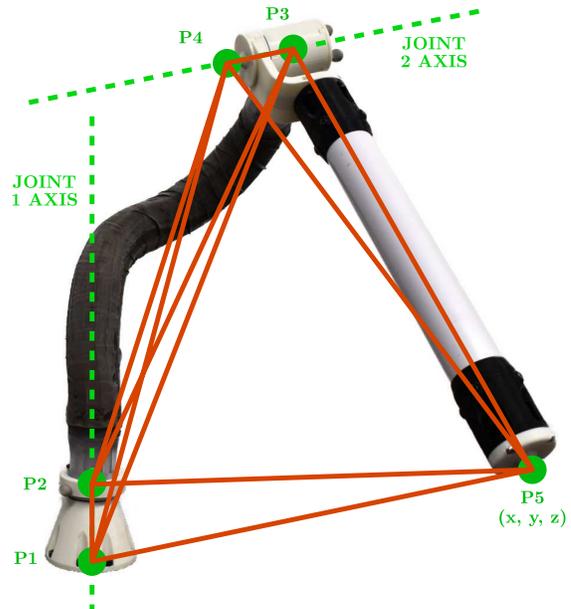}
    \caption{For workspace characterisation, a 2-DOF malleable robot arm can be modelled as the bar-and-joint framework formed by connecting with edges 5 points: $P_1$ and $P_2$, which define the first axis; $P_3$ and $P_4$, which define the second axis; and $P_5$, which corresponds to the centre of the end effector.}
    \label{distancegeometry}
\end{figure}

We define our 2-DOF malleable robot with a vertical rotary joint at the base, connected co-linearly to one end of a malleable link, which terminates at a second rotary joint mounted perpendicularly to the other end of the malleable link. A second, rigid link is then attached to the second rotary joint, also perpendicularly, which then terminates at the end effector. Since a link connecting two skew revolute axes can be modelled as a tetrahedron by taking two points on each of these axes and connecting them all with edges, and a rigid link connected to a revolute axis can be modelled as a triangle by taking two points on the axis and a point at the end of the link and connecting them all with edges \cite{rojas-coupler}, then we can model a 2-DOF malleable robot as the bar-and-joint framework involving 5 vertices (points) and 10 edges shown in Fig.~\ref{distancegeometry}, where $P_5$ corresponds to the centre of the end effector.

Given a sequence of five points $P_1$, $P_2$, $P_3$, $P_4$, $P_5$, the Cayley-Menger determinant of these is defined as \cite{rojas-thesis}
\small
\begin{equation*}
    D(1,2,3,4,5)=-\frac{1}{16}\left| \begin{array}{cccccc}
                0   &   1         & 1       & 1       & 1       & 1       \\
                1   &   0         & s_{1,2} & s_{1,3} & s_{1,4} & s_{1,5} \\
                1   &   s_{1,2}   & 0       & s_{2,3} & s_{2,4} & s_{2,5} \\
                1   &   s_{1,3}   & s_{2,3} & 0       & s_{3,4} & s_{3,5} \\
                1   &   s_{1,4}   & s_{2,4} & s_{3,4} & 0       & s_{4,5} \\
                1   &   s_{1,5}   & s_{2,5} & s_{3,5} & s_{4,5} & 0       \\
                \end{array} \right|,
\end{equation*}
\normalsize
where $s_{i,j} = d^2_{i,j}=||\mathbf{p}_i-\mathbf{p}_j||^2$ is the squared distance between $P_i$ and $P_j$, and $\mathbf{p}_i$ is the position vector of point $P_i$ in the global reference frame. For the general point sequence $P_{1}$, $P_{2}$,$\ldots$,$P_{n}$, the Cayley-Menger determinant gives $(n-1)!^2$ times the squared hypervolume of the simplex spanned by the points in $\mathbb{E}^{n-1}$ \cite{Menger}. Hence, $D(1,2,3,4,5) = 0$ in $\mathbb{E}^3$. Since $D(1,2,3,4,5) = D(4,3,2,1,5) = 0$, using properties of the determinant of block matrices \cite{Powell}, it can be shown that this condition can be compactly expressed using $3\times3$ matrices as
\begin{align}
    D(1,2,3,4,5) = 2\,s_{1,2}\,s_{1,5}\,s_{2,5}\det(\mathbf{A}-\mathbf{B}\mathbf{C}\mathbf{B}^T)=0, \label{eq:determinant2}
\end{align}
where
\begin{align*}
    \mathbf{A} &= \left[ \begin{array}{ccc}
                0   &   1         & 1       \\
                1   &   0         & s_{3,4} \\
                1   &   s_{3,4}   & 0       \\
                \end{array} \right],\, 
    \mathbf{B} = \left[ \begin{array}{ccc}
                1   &   1         & 1       \\
                s_{2,4}   &   s_{1,4}         & s_{4,5} \\
                 s_{2,3}   &   s_{1,3}         & s_{3,5} \\
                \end{array} \right], \textrm{ and} \\           
    \mathbf{C} &= \frac{1}{2}\left[ \begin{array}{ccc}
                -\frac{s_{1,5}}{s_{1,2}\,s_{2,5}}   &  \frac{1}{s_{1,2}}         &    \frac{1}{s_{2,5}}    \\
                \frac{1}{s_{1,2}}   &   -\frac{s_{2,5}}{s_{1,2}\,s_{1,5}}         & \frac{1}{s_{1,5}} \\
                 \frac{1}{s_{2,5}}   &   \frac{1}{s_{1,5}}         & -\frac{s_{1,2}}{s_{1,5}\,s_{2,5}} \\
                \end{array} \right].
\end{align*}

\begin{figure*}[t!]
    \centering
    \includegraphics[width=\textwidth]{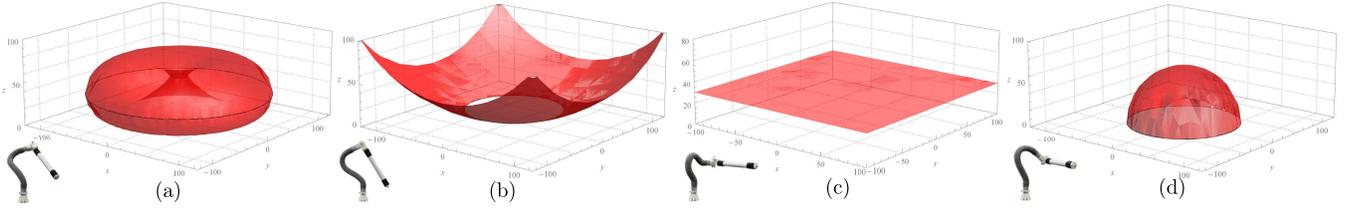}
    \caption{Simulated example workspaces for each type of robot topology achievable by a 2-DOF malleable robot: \textbf{(a)} General Articulated, \textbf{(b)} PUMA-like, \textbf{(c)} SCARA, and \textbf{(d)} Spherical.}
    \label{simulationworkspace}
\end{figure*}

Following the notation of Fig.~\ref{distancegeometry}, equation \eqref{eq:determinant2} is solely satisfied in the points in $\mathbb{E}^{3}$ where a 2-DOF malleable robot can physically exist. This fact can be exploited to compute the Cartesian equation of the robot workspace, say $\Gamma(x, y, z)$, by deriving the locus of point $P_5$, the end effector, whose coordinates are $\mathbf{p}_5 = (x, y, z)$ in a particular reference frame. To simplify this computation we can assume, without loss of generality, that $P_1$ equals the origin of the global reference frame and that $P_2$ is located in the positive side of the $z$-axis, such that $\mathbf{p}_1 = (0, 0, 0)$ and $\mathbf{p}_2 = (0, 0, d_{1,2})$. Therefore,
\begin{align}
\begin{split}
    s_{1,5} &= x^2 + y^2 + z^2\\
    s_{2,5} &= x^2 + y^2 + z^2 - 2d_{1,2}z + s_{1,2}. \label{eq:basedistances}
\end{split}
\end{align}

Substituting equation \eqref{eq:basedistances} into equation \eqref{eq:determinant2}, fully expanding the result and rearranging terms, we get
\begin{align}
    \Gamma(x, y, z) & \myeq  q_0\,(x^2+y^2+z^2)^2+q_1\,d_{1,2}\,z\,(x^2+y^2+z^2) \nonumber \\
    & q_2\,x^2+q_2\,y^2+q_3\,z^2+q_4\,d_{1,2}\,z+q_5, \label{eq:Gamma}
\end{align}
where $q_i,\,i=0,\ldots,5$ are polynomials in $s_{1,2}=d_{1,2}^2$, $s_{1,3}$, $s_{1,4}$, $s_{2,3}$, $s_{2,4}$, $s_{3,4}$, $s_{3,5}$, and $s_{4,5}$. $\Gamma(x, y, z)$ is an algebraic surface of degree 4 (a quartic surface) that corresponds to the workspace surface, traced by the end effector (point $P_5$), of a 2-DOF malleable robot. The expressions of the polynomials $q_i$ cannot be included here due to space limitations; these polynomials can be easily reproduced using a computer algebra system following the steps given above.

By providing constraints to the two revolute axes of the malleable robot, we can define certain workspace categories belonging to specific robot configurations (topologies). Malleable robots are a general purpose serial robot, and so follow similar applications where the task workspace defines the configuration. The robot configurations we define are spherical, PUMA-like, SCARA, and general articulated. The constraints for each of them are discussed next.

\subsection{Spherical (or variable radius) case}
In a spherical robot configuration, the two revolute axes of the robot coincide at the base, such that, according to the notation of Fig.~\ref{distancegeometry}, points $P_1$ and $P_3$ are coincident. Thus, $s_{1,3} = 0$, $s_{2,3} = s_{1,2}$, and $s_{3,4} = s_{1,4}$. Substituting these values into \eqref{eq:Gamma}, we obtain
\begin{align}
    \Gamma_A(x, y, z) \myeq x^2 + y^2 + z^2 - s_{3,5}=0,
\end{align}
which corresponds to the equation of a sphere of radius $d_{3,5}$ centred at $P_1$. Observe that in this case the radius $d_{3,5}$ is not constant, it can be adjusted according to need. An example of this workspace can be seen in Fig.~\ref{simulationworkspace}(d).

\subsection{PUMA-like (or variable centre and radius) case}
In a PUMA-like robot configuration, the two revolute axes of the robot are perpendicular and coincide at a point located in the positive side of the $z$-axis, such that points $P_2$ and $P_4$ are coincident, and the angle $\angle P_1P_2P_3$ is $\frac{\pi}{2}$. Thus, $s_{2,4} = 0$, $s_{1,4} = s_{1,2}$, $s_{3,4} = s_{2,3}$, and $s_{1,3} = s_{1,2}+s_{2,3}$. Substituting these values into \eqref{eq:Gamma}, we get
\begin{align}
    \Gamma_B(x, y, z) \myeq x^2 + y^2 + (z-d_{1,2})^2-s_{4,5}=0,
\end{align}
which corresponds to the equation of a sphere of radius $d_{4,5}$ centred at $P_2$. Observe that in this case the centre $(0, 0, d_{1,2})$ and radius $d_{4,5}$ are not constant, they can be adjusted according to need. The same equation is obtained when the perpendicularity of the two axes is relaxed. An example of such a workspace can be seen in Fig.~\ref{simulationworkspace}(b).

\subsection{SCARA (or planar) case}
In a SCARA robot configuration, the two revolute axes of the robot are parallel. Using projective geometry arguments, this implies that there exist a point in the second axis, say $P_3$, such that the distance between it and the $xy$-plane is $\delta$, with $\delta>0$, $\delta \to \infty$. Hence, $d_{1,3}=\delta$, $d_{2,3}=d_{1,2}+\delta$, $d_{3,4}=z_{4}+\delta$, $d_{3,5}=z_{5}+\delta$, being $z_{i}$ the distance between $P_i$ and the $xy$-plane. Substituting these values into \eqref{eq:Gamma}, we obtain an equation that can be written as a quadratic polynomial in $\delta$, say $\Omega = k_2(x,y,z)\delta^2+k_1(x,y,z)\delta+k_0(x,y,z) = 0$. By factoring out $\delta^2$ in this polynomial, we get $\Omega =\delta^2(k_2(x,y,z)+\frac{k_1(x,y,z)}{\delta}+\frac{k_0(x,y,z)}{\delta})=0$. Since $\delta \to \infty$, then $\Omega = k_2(x,y,z)=0$. 

Since the two revolute axes of the  robot are  parallel, we have to include additional constraints in $\Omega=k_2(x,y,z)=0$, that is, $P_2$=$P_4$=$P^\infty$. This implies that $s_{2, 4} = 0$ and $d_{1, 4} = d_{1, 2}$. Substituting these values into $\Omega=k_2(x,y,z)=0$, we get $(z_4-d_{1, 2})\,s_{1, 2}\,\Phi(x,y,z) = \Phi(x,y,z) = 0$. We can then include the final constraint $z_4 = d_{1, 2}$ (as $P_2$=$P_4$) in the result ($\Phi(x,y,z)$). This yields,
\begin{equation*}
     \left( z-{\it z_5} \right)  \left( {x}^{2}+{y}^{2}+{z}^{2}-2\,d_{{1,2}
}z+{d_{{1,2}}}^{2}-s_{{4,5}} \right) = 0.
\end{equation*}
Following a similar procedure in the above equation to that done for $\delta$, but in this case for $d_{1,2}$ ($d_{1,2} \to \infty$ since $P_2=P_4=P^\infty$), we finally get
\begin{align}
    \Gamma_C(x, y, z) \myeq \left( z-{\it z_5} \right)=0, \label{eq:GammaSCARA}
\end{align}
which corresponds to the equation of a plane parallel to the $xy$-plane. Observe that $z_5$, the distance between the end effector and the $xy$-plane, is not constant and can be adjusted according to need. An example of this workspace can be seen in Fig.~\ref{simulationworkspace}(c).

\subsection{General articulated}
We define the general articulated robot configuration as any robot configuration that does not comply with any of the constraints of the 3 other defined robot configurations, thus the form of the workspace surface in this case is $\Gamma(x, y, z)=0$ (equation \eqref{eq:Gamma}). An example of this workspace, which corresponds to a torus, can be seen in Fig.~\ref{simulationworkspace}(a).

\begin{figure}[t!]
    \centering
    \includegraphics[width=0.9\columnwidth]{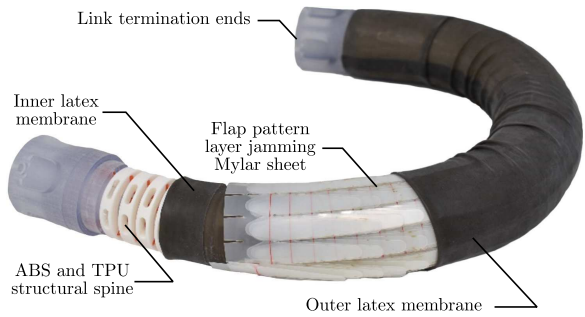}
    \caption{Design of the developed variable stiffness malleable link showing components and overall length/width ratio.}
    \label{malleablelink}
\end{figure}

\section{Malleable Robot Design}

\subsection{Malleable link}
To achieve the variable structure of the malleable link, variable stiffness technology was used, specifically layer jamming with internal structural support \cite{clark-stiffness}, as this allowed for both flexible shaping of the link, as well as rigid fixation of a given configuration. Further, layer jamming has been shown to give the highest jamming stiffness while utilising a very small volume compared to other simple pneumatic jamming solutions \cite{clark-assessing}. Due to the increased size of the malleable link (50mm \diameter) compared to typical variable stiffness links designed for continuum surgery ($\sim$15mm \diameter), other variable stiffness technologies capable of demonstrating higher stiffness, such as shape memory alloys (SMAs) or low melting point alloys (LMPAs) were not considered due to the significant increase in transition time. 

Layer jamming was achieved by flap pattern \cite{kim-design} laser cut Mylar sheet (0.18 mm), of 12 flaps spanning the circumference of the link, with a minimum of 11 overlapping layers always in contact. The flap pattern was contained within two cylinder membranes of latex sheet (0.25 mm), and sealed with link termination ends 3D printed from Vero Clear on a Stratasys Objet 500, which also provided mounting points for an internal structural spine to prevent excessive deformation under extreme bending of the link, as well as mounting points to attach the other components of the robot. The internal spine \cite{clark-stiffness} was 3D printed from Acrylonitrile Butadiene Styrene (ABS), with flexible couplings connecting the spinal segments printed from Ninjaflex material. By changing the pressure inside the sealed latex membranes using a vacuum pump (BACOENG 220V/50Hz BA-1 Standard), the Mylar layers compress together, and the cumulative friction causes a significant increase in rigidity, proportional to the negative pressure applied. The components of the implemented malleable link can be seen in Fig.~\ref{malleablelink}.

\subsection{Joints and rigid link}
\begin{figure}[t!]
    \centering
    \includegraphics[width=0.9\columnwidth]{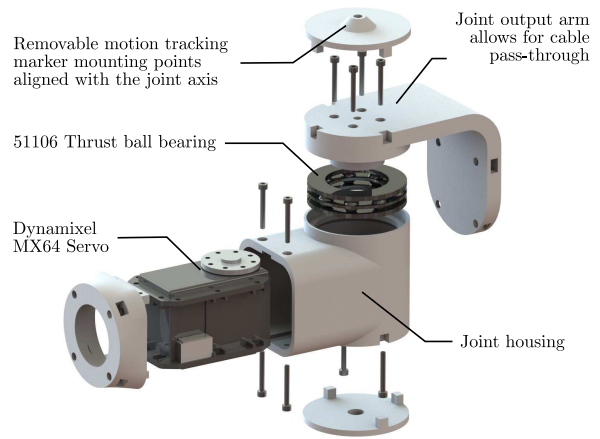}
    \caption{Secondary rotary joint exploded CAD detailing components.}
    \label{malleablejoint}
\end{figure}

The primary joint, positioned at the base of the robot, provides rotation in the z-axis. The secondary joint was positioned at the end of the malleable link, providing rotation in the axis perpendicular to the termination end. Both joints were constructed from a Dynamixel MX-64 servo motor, with a 3D printed ABS housing, and a thrust ball bearing (size 51106) providing force distribution of the motor torque to the output side of the joint. The secondary joint components can be seen in Fig.~\ref{malleablejoint}. The rigid link attached to the secondary joint has a length of 370 mm (actual distance of 450 mm between joint axis and end effector). The link was composed of a 42 mm~$\diameter$ Polypropylene tube, and was attached to the robot using 3D printed ABS link ends similar to those used on the malleable link.

\begin{figure*}[t!]
    \centering
    \includegraphics[width=0.99\textwidth]{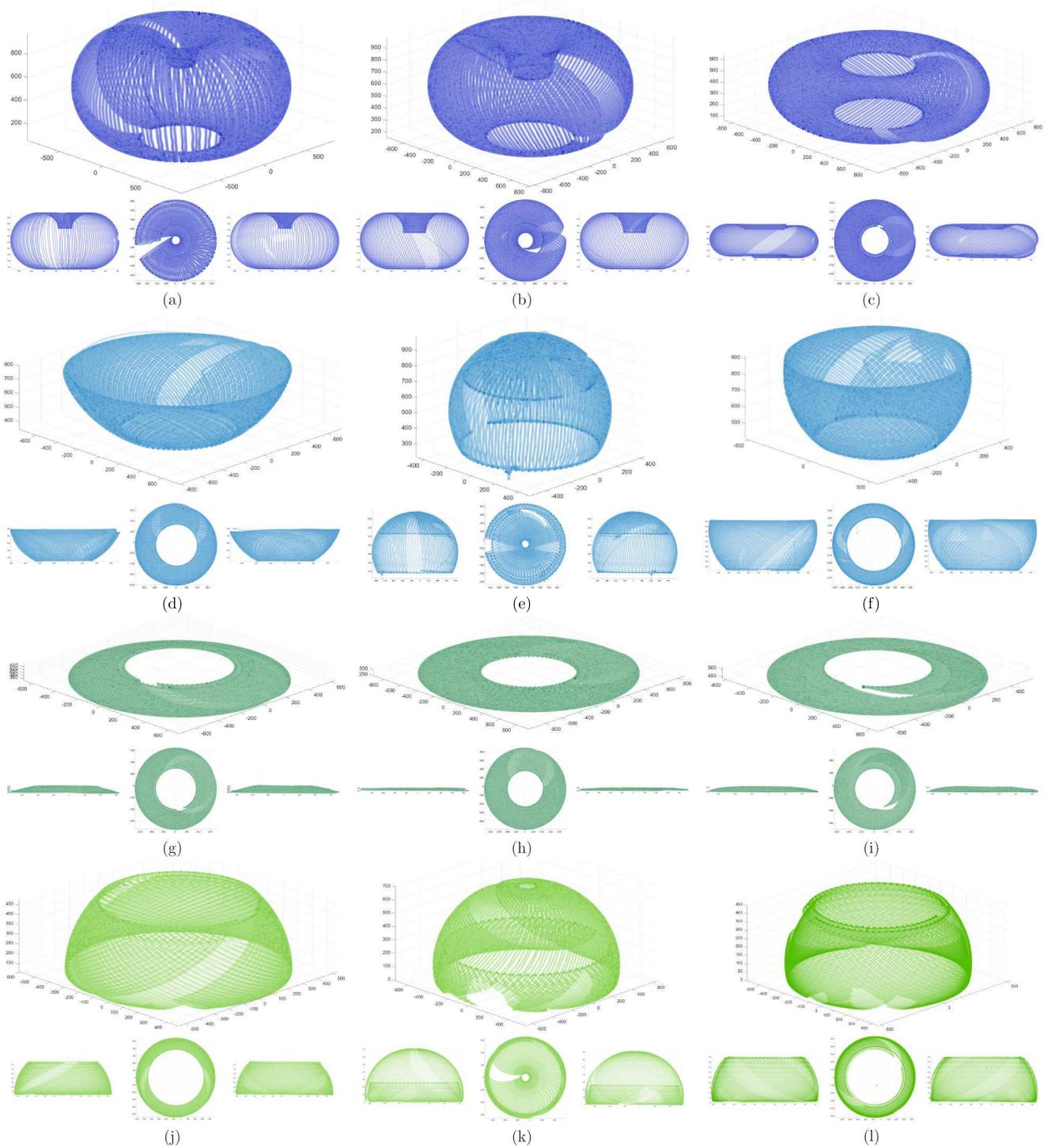}
    \caption{Experimental results showing tracked overall workspaces of the 2-DOF malleable robot in 3 alternative configurations for each type of robot topology, (Top: ISO, Bottom Left to Right: YZ, XY, and XZ): (\textbf{a})(\textbf{b})(\textbf{c}) General Articulated (dark blue), (\textbf{d})(\textbf{e})(\textbf{f}) PUMA-like (light blue), (\textbf{g})(\textbf{h})(\textbf{i}) SCARA (dark green), and (\textbf{j})(\textbf{k})(\textbf{l}) Spherical (light green). Constant distances (in mm) across all configurations were $d_{1,2}$ = 58, $d_{3,4}$ = 49, $d_{3,5}$ = 455, and $d_{4,5}$ = 460. Specific distances (in mm) for each configuration were (\textbf{a}) $d_{1,3}$ = 671, $d_{1,4}$ = 670, $d_{1,5}$ = 1009, $d_{2,3}$ = 626, $d_{2,4}$ = 625, $d_{2,5}$ = 981, (\textbf{b}) $d_{1,3}$ = 668, $d_{1,4}$ = 656, $d_{1,5}$ = 385, $d_{2,3}$ = 620, $d_{2,4}$ = 609, $d_{2,5}$ = 3647, (\textbf{c}) $d_{1,3}$ = 593, $d_{1,4}$ = 618, $d_{1,5}$ = 726, $d_{2,3}$ = 558, $d_{2,4}$ = 581, $d_{2,5}$ = 718, (\textbf{d}) $d_{1,3}$ = 650, $d_{1,4}$ = 667, $d_{1,5}$ = 969, $d_{2,3}$ = 603, $d_{2,4}$ = 619, $d_{2,5}$ = 930, (\textbf{e}) $d_{1,3}$ = 523, $d_{1,4}$ = 525, $d_{1,5}$ = 529, $d_{2,3}$ = 466, $d_{2,4}$ = 469, $d_{2,5}$ = 497, (\textbf{f}) $d_{1,3}$ = 642, $d_{1,4}$ = 669, $d_{1,5}$ = 551, $d_{2,3}$ = 588, $d_{2,4}$ = 614, $d_{2,5}$ = 510, (\textbf{g}) $d_{1,3}$ = 500, $d_{1,4}$ = 544, $d_{1,5}$ = 562, $d_{2,3}$ = 451, $d_{2,4}$ = 494, $d_{2,5}$ = 520, (\textbf{h}) $d_{1,3}$ = 569, $d_{1,4}$ = 596, $d_{1,5}$ = 503, $d_{2,3}$ = 540, $d_{2,4}$ = 564, $d_{2,5}$ = 474, (\textbf{i}) $d_{1,3}$ = 535, $d_{1,4}$ = 581, $d_{1,5}$ = 647, $d_{2,3}$ = 483, $d_{2,4}$ = 528, $d_{2,5}$ = 604, (\textbf{j}) $d_{1,3}$ = 395, $d_{1,4}$ = 444, $d_{1,5}$ = 614, $d_{2,3}$ = 343, $d_{2,4}$ = 392, $d_{2,5}$ = 572, (\textbf{k}) $d_{1,3}$ = 511, $d_{1,4}$ = 559, $d_{1,5}$ = 695, $d_{2,3}$ = 472, $d_{2,4}$ = 520, $d_{2,5}$ = 666, (\textbf{l}) $d_{1,3}$ = 349, $d_{1,4}$ = 396, $d_{1,5}$ = 443, $d_{2,3}$ = 299, $d_{2,4}$ = 347, $d_{2,5}$ = 441.}
    \label{expresults}
\end{figure*}

\begin{figure*}[t!]
    \centering
    \includegraphics[width=\textwidth]{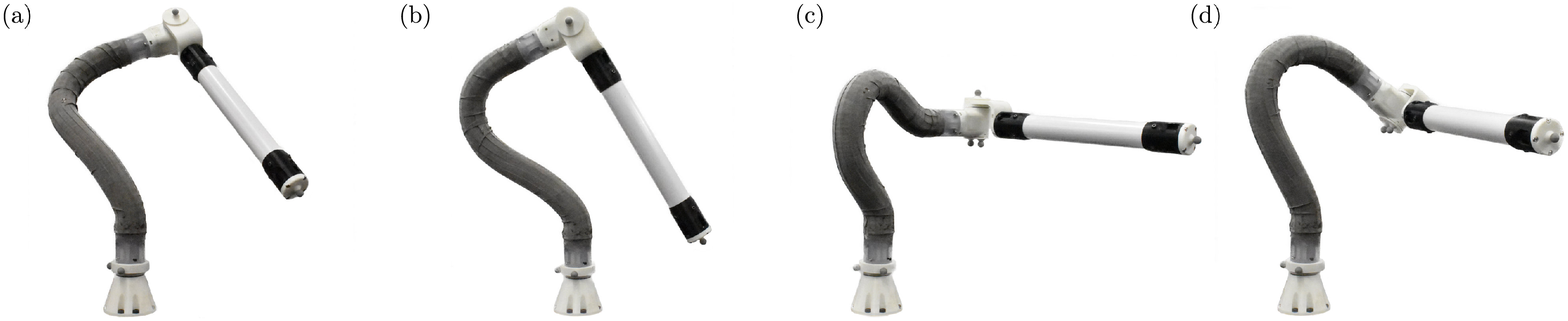}
    \caption{Malleable robot topology configurations for each of the defined workspaces: \textbf{(a)} General Articulated, \textbf{(b)} PUMA-like, \textbf{(c)} SCARA, and \textbf{(d)} Spherical.}
    \label{malleablerobotconfiguration}
\end{figure*}

\section{Performance Evaluation}
To demonstrate the viability of the malleable robot concept and confirm its capability of generating an infinite number of workspaces for different robot topologies, the workspace of the developed malleable robot was measured. Motion tracking markers were attached to each joint of the robot and the end effector. 7 OptiTrack Flex3 cameras were used to track the movement of the robot. The calibration report of the cameras detailed a mean 3D error for overall projection as 0.455mm and overall wand error as 0.081mm. Before each experiment, the desired configuration of the robot was selected. Using live tracking feedback from the motion tracking cameras, the malleable robot was manually shaped and fixed in position, with the pose measured by the tracking marker positions. 12 robot geometries were selected, 3 for each of the workspace configurations defined in section II. 

The geometric parameters of the robot (\emph{i.e.}, distances) for each assessed configuration are presented in Fig.~\ref{expresults}, where the experiment results obtained are shown. Once each geometry was confirmed, the tracking markers on the joints were removed, leaving only the single marker on the end effector. The end effector was then moved throughout the entire workspace of each robot configuration by progressing through all possible joint positions in steps of 0.088$^\circ$. Due to joint limits, the maximum angles actually achievable by each joint were 10$^\circ$ to 350$^\circ$ and 53$^\circ$ to 307$^\circ$, for the primary and secondary joint, respectively, with 0$^\circ$ aligned with the connecting link. Examples of each robot configuration can be seen in Fig.~\ref{malleablerobotconfiguration}.

\section{Discussion}
The results (Fig.~\ref{expresults}) show 4 distinct workspace configurations related to the topologies general articulated, PUMA-like, SCARA, and spherical, with 3 different variations of geometry within the same topology. For the general articulated, PUMA-like, and spherical configurations, the physical shape of the workspace demonstrated significant variation with the change of geometric parameters, producing flattened tori (general articulated case) [Fig.~\ref{expresults}(c)] and variation in the radius of the resulting sphere (PUMA-like and spherical case) [Fig.~\ref{expresults}(d)-(f) and Fig.~\ref{expresults}(j)-(l)], respectively. For the SCARA case, only variations in workspace height (z-axis) and internal radius were observed, due to the fixed length of the rigid link defining the width of planar surface. On each of the resulting workspaces, we can observe a missing slice, corresponding to areas not accessible by the motion tracking cameras.

Considering the construction of the malleable robot, issues arise from the generated SCARA workspace. The SCARA workspace should be a planar surface (equation \eqref{eq:GammaSCARA}), with variation of the end effector only in the x and y direction. From the experimental results however, we observe a slight variation in height across the workspace. As the robot undergoes movement, the centre of mass of the robot changes. Due to tolerances in the joints, specifically in the primary joint, and wrinkles in the latex membrane, a slight variation in height as the robot progresses through the joint positions can be seen. While this variation is minimal it is an aspect that must be considered in the production of malleable robots. To ensure minimal variation of the malleable link during an experiment, the joint speeds were reduced (25RPM), preventing extreme forces under directional changes.

Finally, from the results we can identify that the manual positioning of the robot (extrinsic reconfiguration) was difficult to accurately achieve some of the intended configurations. As shown by Fig.~\ref{expresults}(f), for instance, where a sphere was desired (PUMA-like case) but an overlapping sphere, a form of a minimal torus, was obtained instead. This implies that the second revolute axis did not intersect the first one as required---and then the robot geometry was that of the general articulated case. While the live feedback from the motion tracking markers allowed improved alignment of the joints over pure visual alignment, it did not account for the accuracy limitations of human manipulation. While the overall accuracy achieved was good for our purposes, as shown by the majority of the workspaces in the correct desired form, it is clear that a slight change in reconfiguration can be critical to the resulting workspace accuracy. In fact, for industrial serial arms, accuracy is key, thus the direction of malleable robots towards collaborative robots, where inaccuracies can easily be dynamically corrected, is key.

\section{Conclusions}
In this paper we presented for the first time the development of a full malleable robot. It consists of two revolute joints, a variable stiffness malleable link, and a rigid link connecting the secondary joint to the end effector. By increasing the pressure within the malleable link, the link can be extrinsically (manually) reshaped, and fixed in position by reducing the pressure, forming an alternative configuration of the robot by varying the relative joint orientations. The design of each malleable robot component was discussed, and using distance geometry, we derived the algebraic surface corresponding to the workspace of the robot. Through simulation, we demonstrated the 4 workspace categories achievable by a 2-DOF malleable robot (spherical, PUMA-like, SCARA, and general articulated). Using motion tracking cameras on the robot prototype, we demonstrated the variable workspace capability of a 2-DOF malleable robot, and confirmed the capacity of generating an infinite number of workspaces, categorised into 4 surface area types. While limitations were presented with the extrinsic reconfiguration of the robot in accurately achieving some desired workspaces, most of the intended geometries were correctly obtained. Future work may consider the implementation of integrated motion sensors, or intrinsic control for achieving desired workspaces.

\addtolength{\textheight}{-8.5cm} 


\newpage
\bibliographystyle{IEEEtran}
\bibliography{references.bib}

\end{document}